\documentclass[conference]{IEEEtran}
\IEEEoverridecommandlockouts
\usepackage[T1]{fontenc}

\usepackage{amsmath}
\usepackage{graphicx}
\usepackage[colorlinks=true,linkcolor=blue,urlcolor=blue,citecolor=blue,breaklinks=true]{hyperref}

\def\BibTeX{{\rm B\kern-.05em{\sc i\kern-.025em b}\kern-.08em
    T\kern-.1667em\lower.7ex\hbox{E}\kern-.125emX}}
    
\makeatletter
\newcommand{\linebreakand}{%
  \end{@IEEEauthorhalign}
  \hfill\mbox{}\par
  \mbox{}\hfill\begin{@IEEEauthorhalign}}
\makeatother

\newcommand\blfootnote[1]{%
  \begingroup
  \renewcommand\thefootnote{}\footnote{#1}%
  \addtocounter{footnote}{-1}%
  \endgroup
}

\IEEEaftertitletext{\vspace{-1.5\baselineskip}}
    
\begin{document}

\title{Probabilistic Forecasting of Day-Ahead Electricity Prices and their Volatility with LSTMs \\
\thanks{We gratefully acknowledge funding from the Helmholtz Association via the grant no.~VH-NG-1727, the grant ``Uncertainty Quantification -- From Data to Reliable Knowledge (UQ)'' no.~ZT-I-0029, the Helmholtz Association’s Initiative and Networking Fund through Helmholtz AI and the Deutsche Forschungsgemeinschaft (DFG, German Research Foundation) via grant no.~491111487. 
The authors gratefully acknowledge the computing time granted through JARA on the supercomputer JURECA~\cite{JURECA} at Forschungszentrum J{\"u}lich.}
}

\author{\IEEEauthorblockN{1\textsuperscript{st} Julius~Trebbien}
\IEEEauthorblockA{\textit{Institute of Energy and} \\
\textit{Climate Research (IEK-10)}\\
\textit{Forschungszentrum J\"ulich}\\
J\"ulich, Germany \\
j.trebbien@fz-juelich.de}
\and
\IEEEauthorblockN{2\textsuperscript{nd} Sebastian~P{\"u}tz}
\IEEEauthorblockA{\textit{Institute for Automation and} \\
\textit{Applied Informatics}\\
\textit{Karlsruhe Institute of Technology}\\
Eggenstein-Leopoldshafen, Germany \\
sebastian.puetz@kit.edu}
\and
\IEEEauthorblockN{3\textsuperscript{rd} Benjamin~Sch{\"a}fer}
\IEEEauthorblockA{\textit{Institute for Automation and} \\
\textit{Applied Informatics}\\
\textit{Karlsruhe Institute of Technology}\\
Eggenstein-Leopoldshafen, Germany \\
benjamin.schaefer@kit.edu}
\and
\IEEEauthorblockN{4\textsuperscript{th} Heidi~S.~Nyg{\aa}rd}
\IEEEauthorblockA{\textit{Faculty of Science and Technology}\\
\textit{Norwegian University of Life Sciences}\\
Ås, Norway \\
heidi.nygard@nmbu.no}
\and
\IEEEauthorblockN{5\textsuperscript{th} Leonardo~Rydin~Gorj\~ao}
\IEEEauthorblockA{\textit{Faculty of Science and Technology}\\
\textit{Norwegian University of Life Sciences}\\
Ås, Norway \\
leo.rydin@nmbu.no}
\and
\IEEEauthorblockN{6\textsuperscript{th} Dirk~Witthaut}
\IEEEauthorblockA{\textit{Institute of Energy and} \\
\textit{Climate Research (IEK-10)}\\
\textit{Forschungszentrum J\"ulich}\\
J\"ulich, Germany \\
d.witthaut@fz-juelich.de}
}

\maketitle

\begin{abstract}
Accurate forecasts of electricity prices are crucial for the management of electric power systems and the development of smart applications. European electricity prices have risen substantially and became highly volatile after the Russian invasion of Ukraine, challenging established forecasting methods. Here, we present a Long Short-Term Memory (LSTM) model for the German-Luxembourg day-ahead electricity prices addressing these challenges.  
The recurrent structure of the LSTM allows the model to adapt to trends, while the joint prediction of both mean and standard deviation enables a probabilistic prediction.
Using a physics-inspired approach -- superstatistics -- to derive an explanation for the statistics of prices, we show that the LSTM model faithfully reproduces both prices and their volatility.
\end{abstract}


\vspace*{-1em}
\section{Introduction}
Electricity prices in Europe underwent a substantial increase in the past two years, driven by an energy crisis coupled with the invasion of Ukraine~\cite{osivcka2022european, Vaughan2022, bottcher2023initial}.
The stark dependence of several European countries on Russian gas and oil, and overall inefficiencies in transmission and operation of the electricity markets, have led to a manifold increase of the exchange market prices~\cite{Hauser2021, Pedersen2022}.
This development has serious economic consequences for power-intensive industries~\cite{Ari2022, zakeri2022energy}.

Accurate forecasts of electricity prices are crucial for smart power systems.
For example, demand-side management is improved by accurate forecasting as it allows to anticipate the optimal scheduling of energy consumption and storage~\cite{shinde2018stackelberg, lin2020research}.
However, the increasing uncertainty of electricity prices affects many optimization problems and should thus be reflected in any forecast~\cite{toubeau2018deep}.

Modeling electricity prices has traditionally been implemented with univariate and multivariate time series analysis~\cite{Weron2006}.
Like many other recent data-centric approaches, electricity price forecasting has received a boost from machine learning (ML) approaches~\cite{Lago2021, hewamalage2021recurrent, Jedrzejewski2022, tschora2022electricity, Trebbien2023}.
As ML algorithms do not require any assumptions about economical mechanisms in advance but intrinsically uncover them, they can adjust to a changing environment of energy systems variables to output accurate predictions.
Recurrent machine learning models are particularly suited to adapt to system changes because of the memory effect encoded in their internal states.
Yet, many machine learning applications focus on predicting only the average and forego analyzing the volatility of prices, i.e., they lack an intrinsic uncertainty quantification.

In this work, we present a forecasting model based on Long Short-Term Memory (LSTM) recurrent neural networks~\cite{hochreiter1997long, Li2021} that predicts both the mean and the standard deviation of day-ahead electricity prices.
The model addresses two major challenges of price forecasting discussed above: (1) the rapid and comprehensive changes in the European electricity market market and (2) the increasing volatility of the prices. 
The LSTM makes use of various power-systems data, primarily load and renewable generation forecasts as well as fuel prices.
Focusing on the day-ahead prices in the German-Luxembourg (DE-LU) bidding zone, we show that the LSTM produces accurate forecasts of electricity prices and price volatility.

In addition, we introduce a novel approach to validate the ability of the model to reproduce the statistics and volatility of electricity prices. 
We take inspiration from superstatistics in order to contrast the statistical properties of electricity price time series~\cite{Beck2003, Beck2005} with the LSTM predictions.
This physics-inspired approach links stochastic volatility with physical principles~\cite{Queiros2005}, allowing us to estimate the volatility of prices.


\section{Background}

\subsection{European Electricity Markets}

Electricity markets play a critical role in coordinating generation and demand prior to the actual delivery of electricity.
Since day-ahead markets trade on a short-term basis, they are most important for coordination and serve as the main reference for general price development.
Market participants use power balance forecasts to elaborate optimal trading strategies one day before actual delivery.
The forecasts include demand and renewable generation to name but a few.

On European day-ahead markets, trading is possible until 12:00 of the day prior to delivery.
Most European exchanges are coupled with Single Day-Ahead Coupling (SDAC) to create coupled Market Clearing Prices (MCP) for the participating bidding zones~\cite{monopolkommission}.
Taking into account all bids and offers as well as network constraints between bidding zones, an algorithm computes the MCP and all implicit cross-border trades.

Due to the increasing share of weather-dependent renewable energy sources and their high volatility~\cite{staffell2018increasing}, which have doubled from 2004 to 2021 to 21.8\% of the total European generation~\cite{StatisticsEU}, the electricity market is also becoming highly volatile~\cite{Han2022}. 
Hence, price forecasting is becoming increasingly difficult.
However, it also enables smart grid applications to be profitable when accurate price forecasts are available.

\subsection{Data}\label{sec:data}

For the task of forecasting electricity prices in the day-ahead market, we only include information that would be available to any market participant in the day-ahead market.
In particular, we only include features that would be available by the market closure at 12:00.
As prediction target, we use hourly day-ahead electricity prices for Germany, collected from the ENTSO-E transparency platform~\cite{entso-e_transparency}.
Notably, Germany shares its bidding zone with Luxembourg.

As inputs for our prediction model, we collect power system features from the ENTSO-E transparency platform \cite{entso-e_transparency} and fuel prices from various platforms detailed below.
An overview of all features used is provided in Tab.~\ref{tab:data}. All data has a 1-hour resolution. Power system features include forecasts of load, wind generation, and solar generation which are the main factors driving day-ahead electricity prices~\cite{Trebbien2023}.
The forecast of wind generation is aggregated from on- and offshore generation.

Electricity prices in one bidding zone are affected by neighboring zones due to the SDAC.
To capture these interactions while keeping the feature set light, we include the residual load forecast for each neighbouring bidding zone.
The residual load is the difference of the load and the variable renewable generation. 
Notably, the data on solar generation in Poland (PL) is missing prior to 2020-04-10.
We modelled this time series using a simple linear regression with respect to the solar generation of its neighboring countries.
Furthermore, we did not include Swedish bidding zone 4 (SE4) due to its high amount of missing data points. 

\begin{table}[tb]
\caption{Features and their respective units. Abbreviations: DA: Day-ahead; Nuc. Avail.: Nuclear availability; Res.: Residual$^\dagger$}
\label{tab:data}
\centering
\vspace*{-.5em}
\begin{tabular}{l|l}
\textbf{Feature}         & \textbf{Unit} \\ \hline
DA Load DE-LU            & MW      \\
DA Solar DE-LU           & MW      \\
DA Wind DE-LU            & MW      \\
DA Res. Load AT, BE, CH, CZ      & MW      \\
~~~ DK1, DK2 , FR, NL, NO2, PL      &         \\
Nuc. Avail. DE-LU, FR    & MW      \\ \hline
Gas Price                & EUR/MWh \\
Oil Price                & USD/bbl \\
Coal Price               & USD/t   \\
CO2 Price                & EUR/t       
\end{tabular}
\vspace*{-2em}
\end{table}

Following Ref.~\cite{rinne_radioinactive_2019}, we complement the feature set with the available nuclear capacity.
For each hour, the available capacity is calculated as the installed capacity minus the planned unavailability of nuclear power plants~\cite{entso-e_installed_nodate, entso-e_unavailability_nodate}.
We restrict the data to Germany-Luxembourg and France, as France has by far the largest installed nuclear capacity in Europe.

In addition to power system features, we include several fuel prices in the dataset.
Notably, fuels are traded on markets similar to the stock market, enabling continuous trading during trading hours.
To ensure a realistic forecasting approach, we only take the opening prices for each trading day, shift them by exactly 24 hours, and consider it the fixed price for the entire day.
Gas prices are taken from \cite{investingcom_dutch_2023, marketwatch_ng00_2023}, oil prices are taken from \cite{fred_federal_nodate} and coal prices from \cite{marketwatch_mtfc00_2023}. 
We also include the price of carbon emission certificates in the dataset.
This data was taken from EEX \cite{eex_eex_nodate} for the whole time period.

\blfootnote{$^\dagger$\,Countries according to the ISO-3166 code: DE-LU: Germany-Luxembourg; AT: Austria; CH: Switzerland; CZ: Czechia; DK1 \& DK2: Denmark bidding zones 1 and 2; FR: France; NL: The Netherlands; NO2: Norway bidding zone 2; PL: Poland.}

\vspace*{-1em}\section{Model}
We develop a Long Short-Term Memory (LSTM) model to forecast the German-Luxembourg electricity prices.
LSTMs are a type of recurrent neural network particularly designed to handle long sequences of up to 1000 discrete time steps~\cite{hochreiter1996lstm, hochreiter1997long}.
Using longer time periods as inputs mitigates the extensive search for suitable lagging features.

From the dataset, we use data with a fixed sequence length of 96 hours as input and the electricity price of a single hour as the target.
As the output of the networks, we use two values to create a probabilistic forecast, assuming a normal distribution with the probability density function
\begin{equation}\label{eq:guass}
    \rho(x|\mu,\sigma) = \frac{1}{\sigma \sqrt{2\pi}} e^{-\frac{1}{2} \left ( \frac{x - \mu}{\sigma} \right )^2}.
\end{equation}
The first output value is the mean $\mu$ of the distribution and the second value is the standard deviation $\sigma$. 
We impose a lower bound on the standard deviation, $\sigma \ge 0.01$, to prevent the neural network from predicting negative values.

Due to the dynamical changes in the characteristics of mean and volatility, model performance can decrease after a certain time. 
Therefore, we retrain the model from scratch after one week, using each week as a test set.
Moreover, we disregard any week in which there are less than $120$ hours remaining after the aforementioned removal of missing data points from the evaluation process.
We use $17000$ hours before each test set as train and validation sets, i.e., approximately two years of data.
To ensure the normalization of the data without look-ahead bias, we normalize all data sets using the maximum of each feature in the train set.

The model is optimized using the Adam optimizer~\cite{kingma2014adam} and the Negative Log-Likelihood (NLL) as loss function.
Assuming that the model predicts a normal distribution with mean $\mu_i$ and standard deviation $\sigma_i$, the NLL is given by
\begin{equation}\label{eq:nll}
    \text{NLL}(y, (\mu, \sigma)) = \frac{1}{N} \sum_{i=1}^N  \!\left(\frac{\log (2 \pi \sigma_i^2)}{2} + \frac{(y_i - \mu_i)^2}{2 \sigma_i^2}\right).
\end{equation}
Additionally, the model is evaluated using two metrics for point predictions.
We use the Mean Absolute Error (MAE) and the Symmetric Mean Absolute Percentage Error (SMAPE), given by
\begin{subequations}
\begin{equation}
    \text{MAE}(y, \hat{y}) = \frac{1}{N} \sum_{i=1}^N \left | y_i - \hat{y}_i \right |, \label{eq:mae}
\end{equation}
\begin{equation}
    \text{SMAPE}(y, \hat{y}) = \frac{100}{N} \sum_{i=1}^N \frac{\left | y_i - \hat{y}_i \right |}{\frac{1}{2} \left ( \left |y_i \right | + \left |\hat{y}_i \right | \right )}, \label{eq:smape}
\end{equation}
\end{subequations}
where $y$ is the true value and $\hat{y}$ is the predicted value.

Limiting hyperparameter search to a reasonable degree, we use only a coarse grid search for depth, width, and early stopping to find a well-performing model. 
Dropout is a common regularization technique used in neural networks which randomly removes connections in the networks~\cite{srivastava2014dropout}.
We use a constant dropout of $0.2$ ($20\%$) on each layer.

More details on the model development and a detailed analysis of the role of hyperparameters are provided in Ref.~\cite{trebbien2023thesis}.

\begin{figure*}[t]
    \centering
    \includegraphics[width=0.85\textwidth]{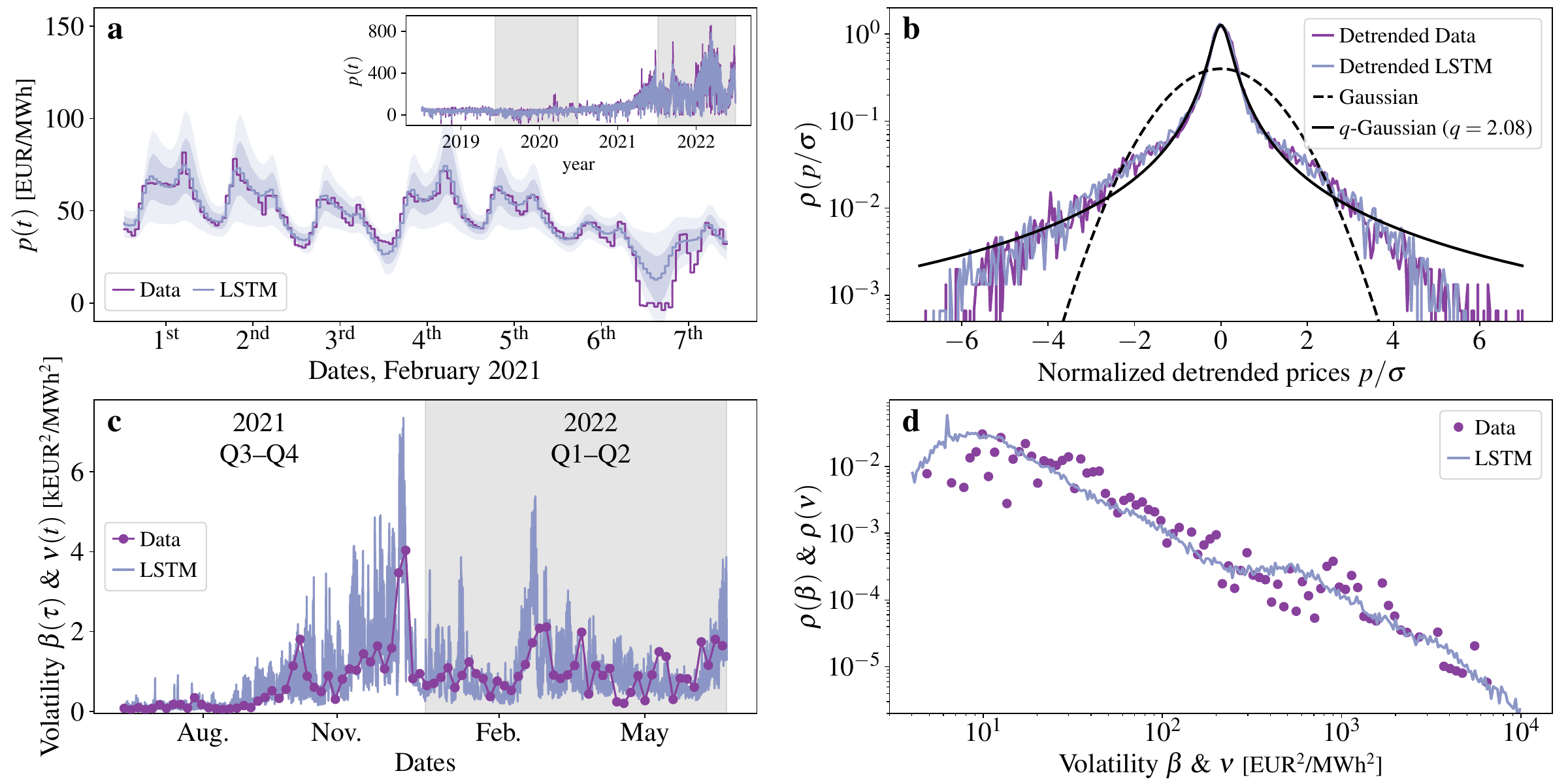}
    \vspace*{-0.5em}\caption{
    The day-ahead electricity prices and the LSTM-probabilistic forecast.
    (\textbf{a}) Prices in February 2021 (4 years in inset), comparing real prices predicted prices (shaded areas indicate 1 and 2 standard deviations from the mean).
    (\textbf{b}) Probability density of the prices and the predicted mean (normalized by standard deviations $\sigma$), with Gaussian and $q$-Gaussian fits.
    (\textbf{c}) Volatility $\beta(\tau)$, estimated from a physics-driven superstatistical approach, and volatility $\nu(t) = 1/\sigma(t)^2$ obtained from the LSTM model.
    (\textbf{d}) Probability density $\rho$ of volatilities $\beta(\tau)$ and $\nu(t)$.
    }\label{fig}\vspace*{-1.5em}
\end{figure*}

\section{Statistical properties of electricity prices and their volatility}
In order to understand volatility in electricity prices, we need to first examine their statistical properties.
Foremost, electricity prices exhibit distributions with heavy tails~\cite{Han2022}.
Moreover, prices are usually correlated processes marked by occasional extreme events~\cite{Weron2006}.
Following Han \textit{et al.}~\cite{Han2022}, we argue that electricity price statistics exhibit a time-scale separation.
At short time scales of up to 4 days (96 hours), day-ahead prices are Gaussian distributed (i.e., they are symmetric and mesokurtic).
Therein, at this scale, they obey Eq.~\eqref{eq:guass}.
This also educates our choice of a 96-hour window for the LSTM implementation.
At long time scales, much longer than 4 days, volatility of the prices induces small and large deviations of the local standard deviation of the process, resulting in a strong leptokurtic distribution of the prices. 
We precise the volatility $\beta(\tau)$ as the inverse local variance of the process $\beta(\tau) = 1/2\sigma(\tau)^2$, wherein we consider volatility in a time scale $\tau\approx 96\,$h.
Hence, we can write an explicit form of the distribution of the day-ahead electricity prices following a superstatistical principle~\cite{Beck2003, Beck2005, Xu2016, Han2022}
\begin{equation}\label{eq:ss}
   \rho(p|\mu) = \int\limits_0^\infty f(\beta) \frac{\sqrt{\beta}}{\sqrt{\pi}} e^{-\beta \left( p - \mu \right)^2} \mathrm{d}\beta,
\end{equation}
wherein some common choices for the distribution of the volatility $f(\beta)$ are known from literature~\cite{Beck2003, Beck2005}.
Importantly, for $f(\beta)$ a Gamma distribution, we obtained a $q$-Gaussian distribution $G_{q,\beta,\mu}(p)$ of the prices
\begin{equation}\label{eq:q_gaussian}
    G_{q,\beta,\mu}(p) = \frac{\sqrt{\beta}}{N_q} e_q(-\beta \left(p-\mu\right)^2),
\end{equation}
with $e_q(x) = [1+(1-q)x]^{1/1-q}$ and $N_q$ a normalization constant~\cite{Han2022}, which play a central role in the analysis of stock market price distribution~\cite{AlonsoMarroquin2019}.

Superstatistical theory allows us to estimate the local volatility under the aforementioned time-scale separation.
This is described in detail in Refs.~\cite{Beck2005, Han2022}. 
Similarly, the LSTM-probabilistic approach yields the standard deviation (or variance) as a function of time.
To distinguish them clearly, we denote the volatility estimated from the superstatistical approach $\beta(\tau)$ (having a 96-hour resolution) and the volatility obtained from the LSTM model $\nu(t)$ (having a 1-hour resolution).
We should note that estimation of superstatistical models is currently only developed for one varying parameter (the volatility $\beta(\tau)$ in our case).
Thus, as in Ref.~\cite{Han2022}, we detrend the prices and predictions using Empirical Mode Decomposition~\cite{Huang1998} by removing the first 5 slowest modes in the data (see Fig.~\ref{fig}, cf. Ref.~\cite{Han2022}).
We will now investigate the ability of the LSTM-probabilistic forecast to accurately reproduce the statistical properties of the German-Luxembourg day-ahead electricity prices.

\section{Results}\label{sec:res}

To establish a comparable model to recent works~\cite{Lago2021, tschora2022electricity}, we conducted a hyperparameter search for the best model with respect to the MAE.
To reduce overall computation time, we evaluated only one year of data for the hyperparameter search.
Specifically, we chose 2021 because it includes prices before and after the start of the European energy crisis.
The best model had a depth of 2, a width of 32, and an early stopping parameter of 200.

The developed LSTM model is able to forecast day-ahead electricity prices with state-of-the-art performance.
Different performance metrics have been evaluated for the 4 years we examined and are summarized in Tab.~\ref{tab:perf}.
We find that the model performance is comparable, if not better than in other recent works. For instance, Tschora \textit{et al.} report an MAE of $7.66$\,EUR/MWh for the test period 2020--2021~\cite{tschora2022electricity}, and demonstrate superior performance compared to established reference models~\cite{Lago2021}.
In comparison, our model yields a lower MAE of $7.08$\,EUR/MWh for the same test period.
We note that this performance is reached despite the fact that our model is not trained to yield the best point forecast, but to minimize the NLL.
Furthermore, we find that the developed model is able to rapidly adapt to the overall pattern and trend of the price time series shown in Fig.~\ref{fig}\textbf{a}.
Furthermore, the predictions closely follow the daily pattern and capture the price dynamics throughout the week. 

\begin{table}[tb]
\centering
\caption{Yearly performance of the LSTM model. 
For probabilistic forecasting, the negative log-likelihood (NLL, Eq.~\ref{eq:nll}) is used. 
For point prediction, the model is evaluated by the mean absolute error (MAE, Eq.~\ref{eq:mae}) and the symmetric mean absolute percentage error (SMAPE, Eq.~\ref{eq:smape}) of the predicted mean price value}
\label{tab:perf}
\vspace*{-.5em}
\begin{tabular}{l|lll}
      & NLL  &  MAE  & SMAPE \\ \hline
2019  & 2.94 & 3.73  & 15.12 \\
2020  & 2.97 & 3.93  & 20.71 \\
2021  & 3.83 & 10.32 & 15.41 \\
2022  & 5.01 & 29.85 & 18.21 \\
all   & 3.69 & 11.92 & 17.42 
\end{tabular}
\vspace*{-2em}
\end{table}

Given our interest in ensuring the model can faithfully reproduce the statistical properties of the price time series, we examine the probability density of the prices, the volatilities, and their distributions.
In Fig.~\ref{fig}\textbf{b}, we show the probability density $\rho$ of the normalized prices and normalized predicted means, along with the best fits for a Gaussian and a $q$-Gaussian distribution.
The $q$-Gaussian distribution, motivated by the superstatistical model presented, agrees well with the distribution of the prices.
Not unexpectedly, Gaussian distributions cannot capture the long tails of the electricity prices.
The superstatistical approach also provides a data-driven method to estimate the volatility $\beta(\tau)$ of the prices, following Ref.~\cite{Han2022}.
In Fig.~\ref{fig}\textbf{c} we display a 1-year snippet of the volatility $\beta(\tau)$ and in Fig.~\ref{fig}\textbf{d} we display the distribution of the volatilities.
We similarly show the LSTM predictions of the volatility $\nu(t)$, which have a 1-hour resolution in contrast with the volatility $\beta(\tau)$ from superstatistics, which have a 96-hour resolution.
Overall, we obtain a good match between the physical principles ruling the price dynamics and the LSTM predictions for both the price as well as the standard deviation.
Obtaining accurate forecasts of both price and standard deviation is crucial for effective uses in, e.g., cost-minimizing demand-side management, battery storage systems, and for power-intensive industries.

\section{Conclusion}
In this work, we tested the application of probabilistic forecasting using an LSTM model for electricity price forecasting.
We developed a simple, two-output LSTM model that predicts the mean and standard deviation of the German-Luxembourg day-ahead electricity prices.
Our interest was to understand the ability of the LSTM in reproducing key statistical properties of time series.
To juxtapose the LSTM forecast, we used a superstatistical approach to recover the statistics and the volatility (which has an inverse relation to the local standard deviation) of the prices.
We show that the LSTM model faithfully forecasts the prices, but moreover, also yields an accurate prediction of the local standard deviation.
Contrasted with the superstatistical approach, we observe that the LSTM can reproduce the correct statistics of the prices, hence, correctly capturing the dynamics of electricity prices.

Machine learning approaches, such as the one presented here, can become a crucial element in the energy markets.
At large, country-wide-scale, they can aid in the forecasting of all relevant power and energy system variables.
At smaller, potentially decentralized, smart- and/or micro-grid applications, pre-trained neural networks can deal with online data to produce valuable predictions that guide the operation of a grid.
Machine learning models can be included in smart grid devices as inexpensive forecasting tools.
A careful validation of the models, including advanced statistical characteristics, is important to assert quality and reliability of the forecasts~\cite{cramer2022validation}.



\end{document}